\documentclass{egpubl}
\usepackage{eg2020}
 
\ShortPresentation      
\usepackage[T1]{fontenc}
\usepackage{dfadobe}  

\usepackage{cite}  
\BibtexOrBiblatex
\electronicVersion
\PrintedOrElectronic
\ifpdf \usepackage[pdftex]{graphicx} \pdfcompresslevel=9
\else \usepackage[dvips]{graphicx} \fi

\usepackage{egweblnk} 

\title{
Frequency-Aware Reconstruction of Fluid Simulations \\with Generative Networks
}

\author[S. Biland et al.]
{\parbox{\textwidth}{\centering
Simon Biland, Vinicius C. Azevedo, Byungsoo Kim and Barbara Solenthaler}
\\
{\parbox{\textwidth}{
\centering
ETH Z{\"u}rich
}}}

\usepackage{amsmath}

\usepackage{graphicx}
\usepackage[footnotesize,sl,SL,hang,tight]{subfigure}
\usepackage{multirow}
\usepackage{wrapfig}
\usepackage{xcolor}

\renewcommand{\vec}[1]{\mathbf{#1}}

\newcommand{\todo}[1]{}

\newcommand\hide[1]{}

\newcommand{\cross}{\times}

\include{amsmath}

\usepackage{amsmath,amsfonts,bm}

\def\eqref#1{equation~\ref{#1}}

\def\1{\bm{1}}

\DeclareMathAlphabet{\mathsfit}{\encodingdefault}{\sfdefault}{m}{sl}
\SetMathAlphabet{\mathsfit}{bold}{\encodingdefault}{\sfdefault}{bx}{n}

\begin{document}


\maketitle
\begin{abstract}
Convolutional neural networks were recently employed to fully reconstruct fluid simulation data from a set of reduced parameters. However, since (de-)convolutions traditionally trained with supervised $\ell_1$-loss functions do not discriminate between low and high frequencies in the data, the error is not minimized efficiently for higher bands. This directly correlates with the quality of the perceived results, since missing high frequency details are easily noticeable. In this paper, we analyze the reconstruction quality of generative networks and present a frequency-aware loss function that is able to focus on specific bands of the dataset during training time. We show that our approach improves reconstruction quality of fluid simulation data in mid-frequency bands, yielding perceptually better results while requiring comparable training time.

\ccsdesc[500]{Computing methodologies~Physical simulation}
\ccsdesc[500]{Computing methodologies~Neural networks}

\printccsdesc
\end{abstract}

\section{Introduction}
\label{sec:Introduction}

Recent work has demonstrated the effectiveness of using neural networks for fluid simulations. Among other work, convolutional neural networks (CNNs) have been used to improve the computational efficiency by either replacing the pressure computation~\cite{Tompson2016} or predicting the fluid motion with an LSTM network~\cite{Wiewel2018}.
A generative deep neural network for parameterized fluid simulations was introduced that reconstructs the velocity field based on a few input parameters~\cite{Kim2019}. This can be used for data compression or to generate simulations by interpolating in the parameter space. 
A main limitation of this previous approach (and related neural fluid solvers) is the difficulty to reconstruct high frequency flow structures, which can be accounted to the $\ell_1$-loss function that is used on the velocity and its gradient. 

Loss functions and their influence on reconstruction quality have been studied in the image processing literature. 
Natural images are known to have an expected power spectrum following an inverse power law: the magnitudes of spectral decompositions are mostly concentrated in lower frequencies. However, high-frequencies with lower magnitudes 
tend to concentrate perceptually important details. Thus, using a mean square error loss will often be more effective in matching lower frequencies that have higher magnitudes and dominate the power spectrum~\cite{Wang2009}. 


Rahaman et al.~\cite{Rahaman2018} used Fourier analysis to investigate which frequencies neural networks tend to learn and reported a phenomenon they call the spectral bias when experimenting with ReLU networks. The authors found that neural networks favor low frequencies, even though they are theoretically universal approximators~\cite{Hornik1991}. Other work has arrived at similar conclusions~\cite{Xu2018,Xu2019}, reporting that dominant low-frequency components are captured quickly during training and high-frequency ones only slowly thereafter (F-Principle). It is argued that this acts as an intrinsic regularizer and filter for noisy input data. 
Complementary losses were used to improve high-frequency reconstruction, such as perceptual losses~\cite{Johnson2016} and adversarial losses (GAN)~\cite{Goodfellow2014,Berthelot2017,Karras2017}. 

Previous perceptual-based loss functions were matched by selectively focusing on frequencies through a weighted wavelet per-band loss combined with a term that penalizes low magnitude values~\cite{Huang2017}. 
The wavelet domain was also employed in several image super resolution approaches. 
A CNN to predict the missing details of wavelet coefficients (sub-bands) of the LR images was used in~\cite{Guo2017}, and a loop architecture to better explore statistical relationships among wavelet coefficients in different bands was applied in~\cite{Geng2018}. Previous work also used clique convolutional blocks~\cite{Zhong2018}, and scatter maps were employed as input to standard CNNs~\cite{Gao2016}.
Wavelet super resolution approaches were also augmented with adversarial losses to better reconstruct the distribution of frequencies~\cite{Huang2019,Zhang2019}. 

Using neural networks for fluid simulations is largely unexplored, and the input data is intrinsically different from images. Our work therefore aims at better understanding the characteristics and reconstruction properties of fluid simulation data. We use the generative approach of~\cite{Kim2019} as a baseline, and present a frequency-aware loss function using Fourier transform. Our results show that reconstruction quality can be improved by considering the different frequency bands in the optimization.

\section{Generative Network for Fluid Flow Reconstruction}


Our work is based on the generative network for fluid simulations presented by Kim et al. {\shortcite{Kim2019}}. Their generator $G$ reconstructs a velocity field $\vec{u}_\vec{c} \in \mathbb{R}^{H \cross W \cross D \cross V_{dim}}$ (height $H$, width $W$, depth $D$, dimension of velocity field $V_{dim}$) from a small set of parameters $\vec{c} = \left [ c_1, c_2, \ldots, c_n \right ] \in \mathbb{R}^n$, which can be represented as a function $G(\vec{c}): \mathbb{R}^n \mapsto \mathbb{R}^{H \cross W \cross D \cross V_{dim}}$. The values $c_i$ parameterize the scene that was used to create the training data. As an example, $c_1$ could be the position and $c_2$ the width of a smoke source, while $c_3$ represents the simulation time step. The network is trained with the baseline loss $L_{b}$ that includes $\ell_1$-losses on the velocity field and its gradient: 
\hspace{-2mm}
\begin{equation} \label{eq:df}
L_{b} = \lambda_{\vec{u}}\left\lVert \vec{u} - \hat{\vec{u}} \right\lVert_1 + \lambda_{\vec{\nabla u}}\left\lVert \nabla \vec{u} - \nabla \hat{\vec{u}} \right\lVert_1,
\end{equation}
where $\vec{u}$ is the simulated ground truth velocity field, $\hat{\vec{u}}=G(\vec{c})$ and $\nabla \hat{\vec{u}}=\nabla G(\vec{c})$, and $\lambda_{\vec{u}}$ and $\lambda_{\nabla\vec{u}}$ serve as weighting factors between the two loss terms. A divergence-free velocity field can be enforced by changing the generator to $G(\vec{c}) = \nabla \cross G'(\vec{c})$, which changes the network to learn a stream function $\psi_\vec{c}$ ($G_{dim}=1$ for 2D and $G_{dim}=3$ for 3D).

\subsection{Analysis of the Reconstruction Quality}
\begin{figure}[h]
	\centering
	\includegraphics[width=0.23\textwidth]{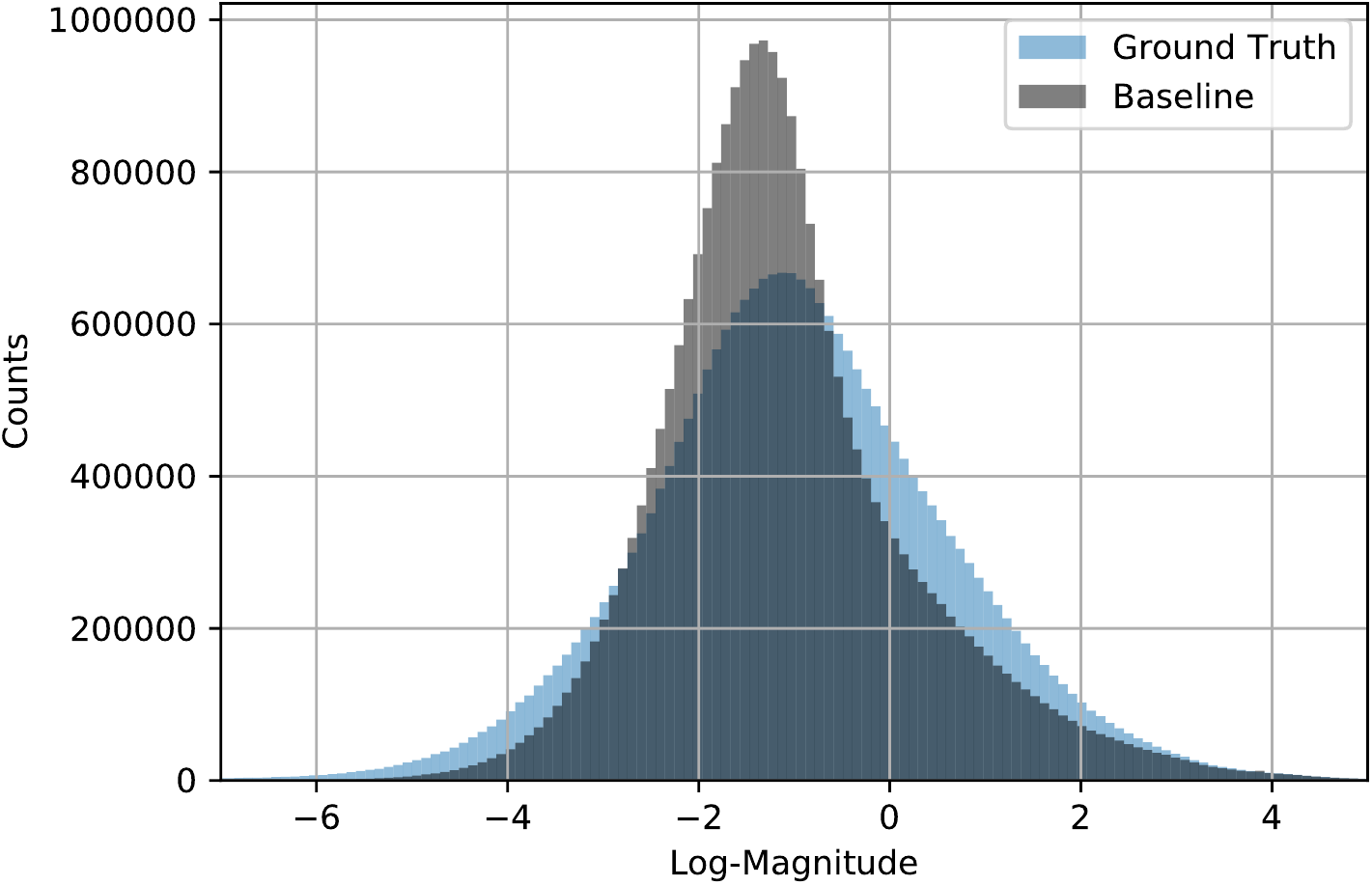}
	\includegraphics[width=0.23\textwidth]{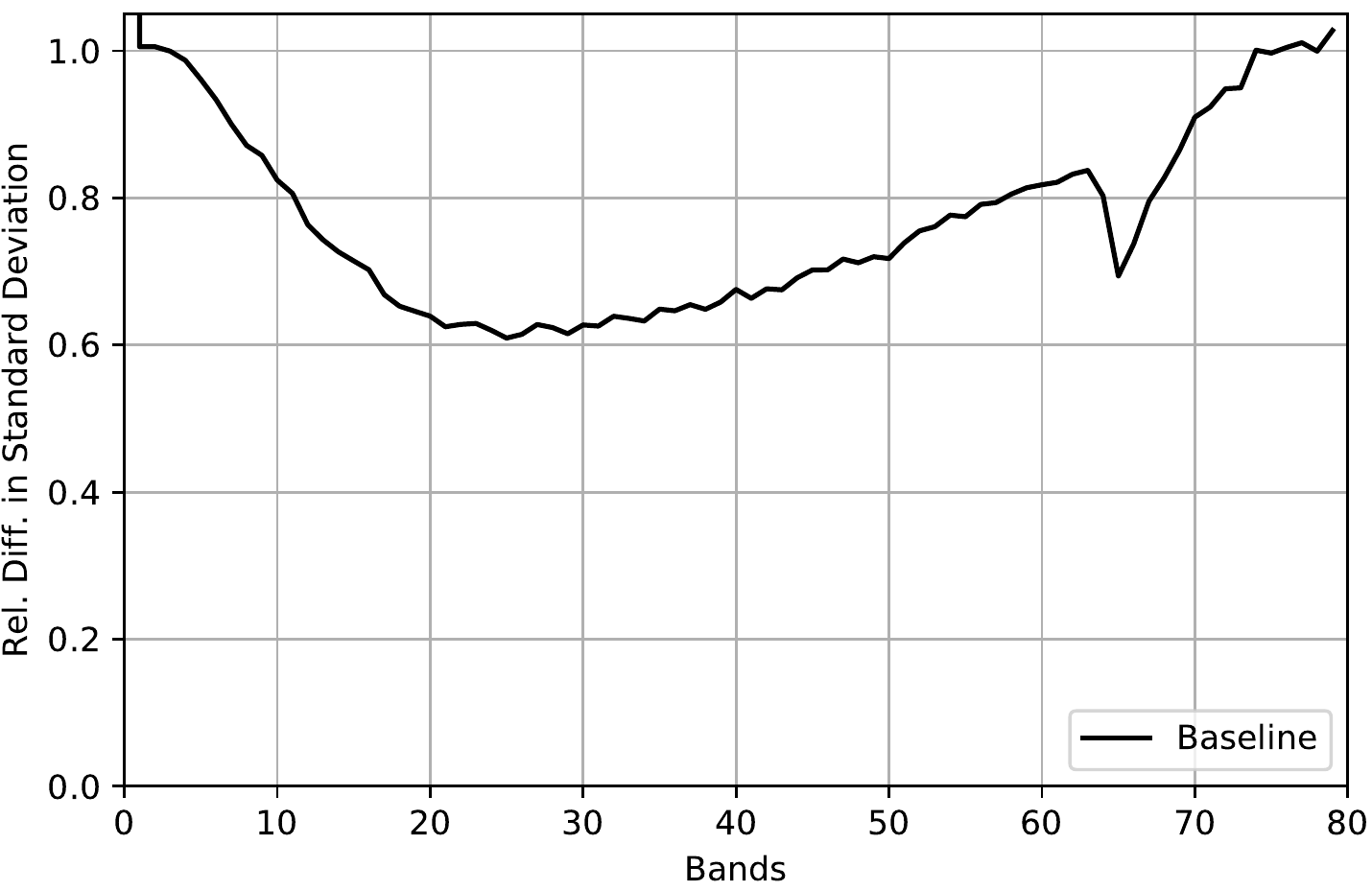}	
	\caption{Left: Distribution of log-magnitudes of the Fourier transform of ground truth and reconstruction. Right: Relative standard deviation of the magnitude of the reconstructed values.\label{fig:distr_error}}
\end{figure}

Although the generative network reliably reconstructs the global structure of the flow, it is apparent that small-scale flow details are lost. 
To verify and analyze this observation quantitatively, we first transform 100 ground truth and reconstructed velocity field samples into the Fourier domain and plot the distribution of the log-magnitude values. Figure~\ref{fig:distr_error} (left) shows that the the distribution of the reconstruction has smaller variance than the ground truth, which can be attributed to the $\ell_1$-loss: to achieve a low error over a large number of training examples, the network strives for universally appropriate values and therefore settles for magnitudes close to the mean of the ground truth data. This property might not be a problem for low frequencies since their magnitudes generally stay within a small interval, but higher frequencies usually have high variance.
This assumption is supported by Figure~\ref{fig:distr_error} (right), which shows the standard deviation of the reconstructed magnitudes separated by bands relative to the standard deviation of the ground truth. Between bands 20 and 30 we observe that the reconstructed standard deviation is only around 60\% of the true standard deviation, which results in the observed clustering around the mean value.

\subsection{Frequency-Aware Loss Function}

\begin{figure}
    \centering
    \includegraphics[width=0.925\linewidth]{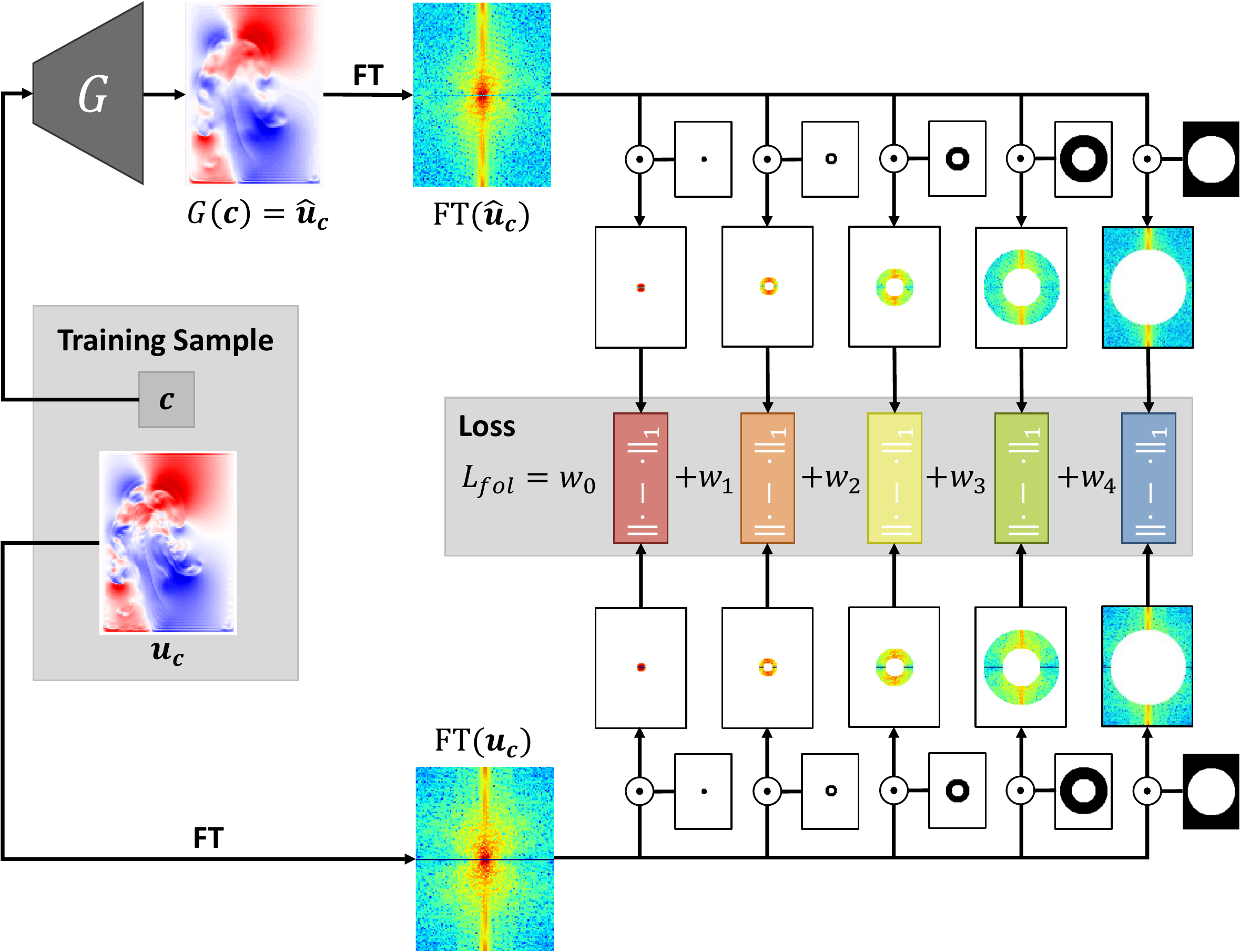}
    \caption{Illustration of the network architecture using a Fourier norm loss with circular bands for fluid flow reconstruction.} 
      \label{fig:architecture}
      \vspace{-40pt}
\end{figure}

Based on the observations discussed above, we propose a frequency-aware method that learns individual frequency bands separately as illustrated in Figure~\ref{fig:architecture}. Both the ground truth $\vec{u}_c$ and the output of the generator $\hat{\vec{u}}_\vec{c}$ are transformed to the Fourier domain ($FT(\vec{u}), FT(\vec{\hat{u}})$) and split into different bands. Then, we compute the $\ell_1$-norm of the difference of each band and aggregate them into a weighted sum.
We compute the error directly in the Fourier domain to avoid expensive computations of inverse transforms and gradients. We define the total loss as $L = \lambda_{o}L_{b}+L_{fol}$, with the Fourier loss $L_{fol}$ given as 
\begin{equation}
    L_{fol} = \sum_b w_b \lVert |FT(\vec{u})_b - FT(\vec{\hat{u}})_b| \rVert.
\end{equation}
$FT(\vec{u})_b$ is the Fourier transform of $\vec{u}$ filtered by band $b$, $|\cdot|$ is the element-wise complex norm, and $\lVert \cdot \rVert_1$ is the $\ell_1$-norm averaged over all pixels in the image. 
%
%
Note that the loss could also be defined on the phase (as in~\cite{Meyer2018}) or magnitude, but we observed low convergence of the phase loss and argue that magnitude loss does not provide enough information for accurate reconstructions.

\begin{figure}[h!]
	\centering
	\newcommand*{\width}{0.24}
		\includegraphics[width=\width\linewidth, trim= 12 20 12 20, clip]{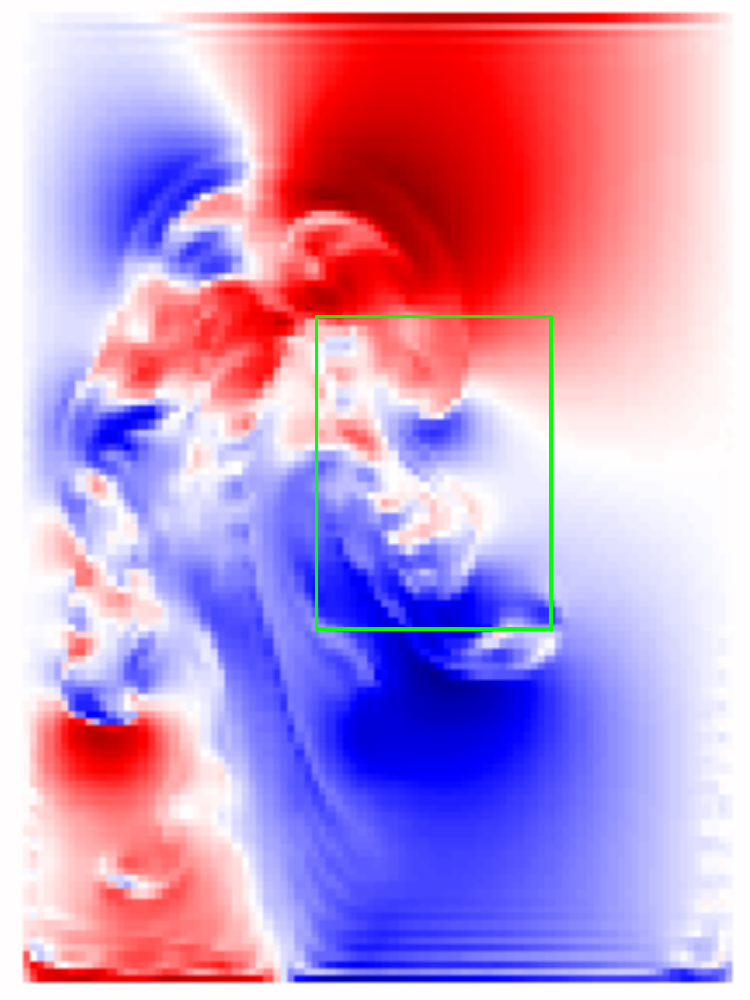}
		\includegraphics[width=\width\linewidth, trim= 12 20 12 20, clip]{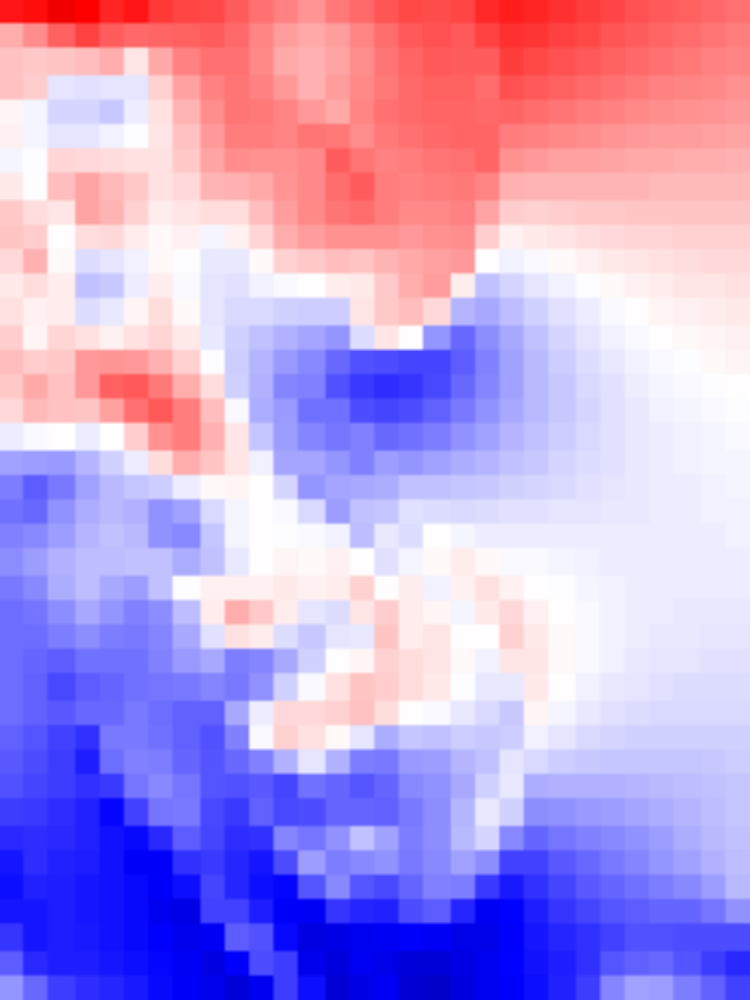}
	\includegraphics[width=\width\linewidth, trim= 12 20 12 20, clip]{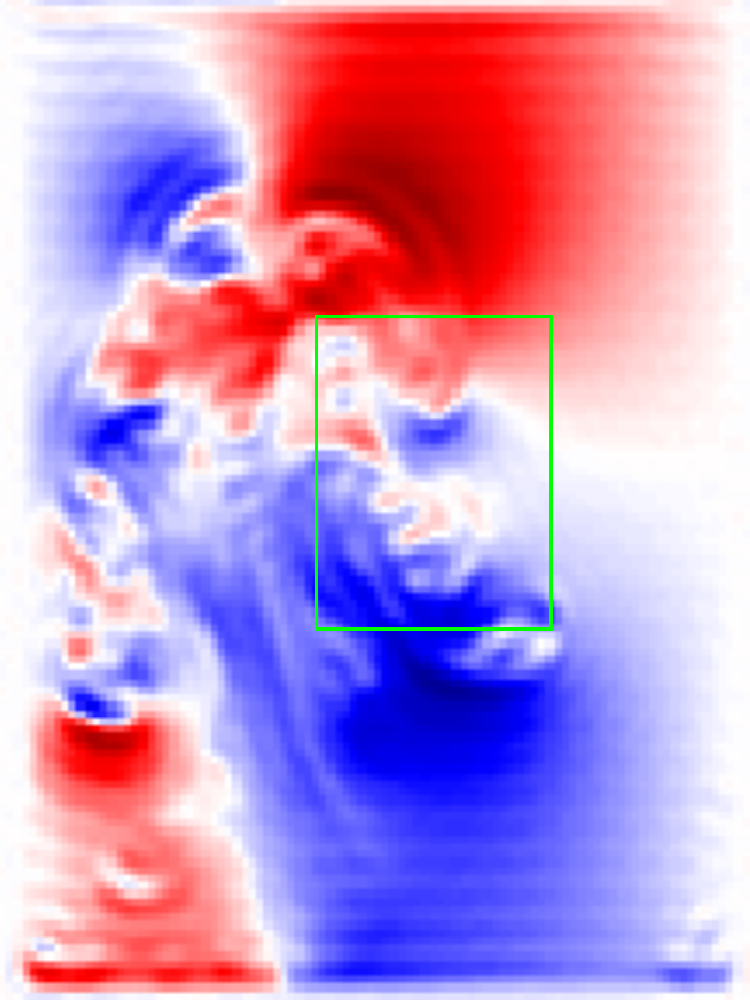}
	\includegraphics[width=\width\linewidth, trim= 12 20 12 20, clip]{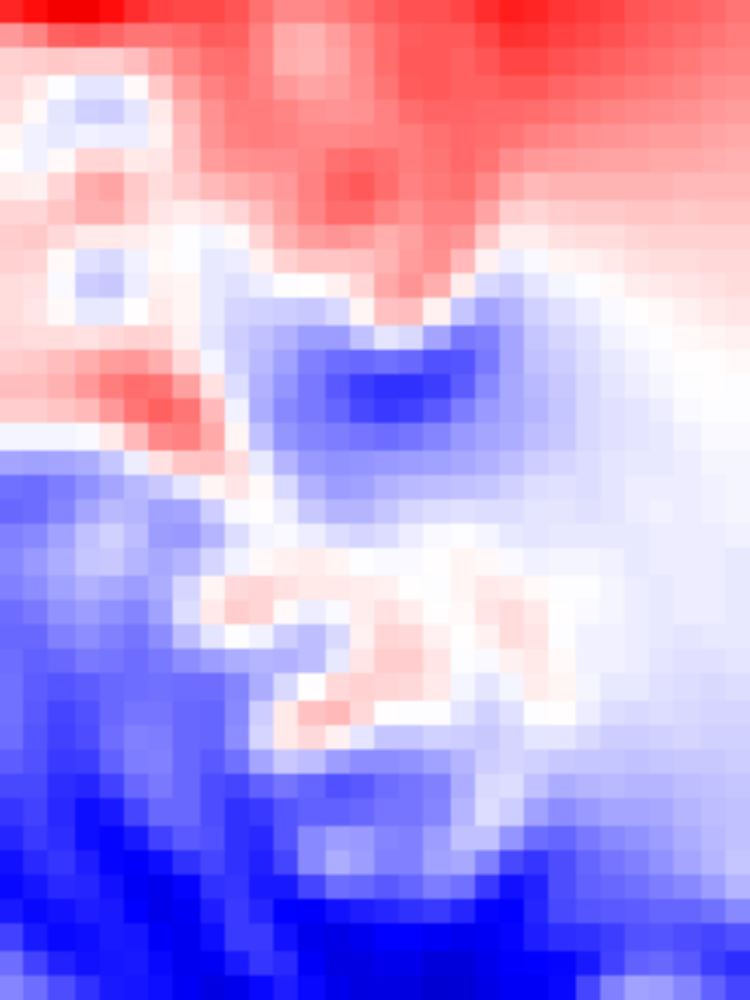}
	\caption{Ground truth (left) and filtered data up to band 30 (right).
	\label{fig:reconstruction_quality}}
	\vspace{-30pt}
\end{figure}

Defining weights $w_b$ is a non-trivial process, as we cannot simply assign higher weights to higher bands. We consider the following observations: 
1) Low frequency bands are visually very important as they define the global flow structure, and we want to retain the high reconstruction quality of the baseline approach in these bands. 
2) Very high frequency bands consist of much noise, which negatively affects the overall convergence of the optimization. 
3) Bands higher than 30 are visually less relevant, as can be seen in Figure~\ref{fig:reconstruction_quality}. 
We therefore reduce the influence of the higher bands, following the intuition that this will eventually lead to a better reconstruction of mid-level frequencies (around bands 10-30). We use a simple heuristic that shifts the weight from the highest frequencies towards lower ones, referred as shift-towards-low (STL). The weight $w_b=\gamma p_b/n$ is inversely proportional to the number of pixels $p$ the band $b$ is covering, with $n$ being the total number of pixels and $\gamma=2$ for all but the highest band where $\gamma=0.5$.

\section{Evaluation and Results}
\label{sec:Results}

\subsection{MRE and Parameter Search}
\begin{figure}[h]
	\centering
	\includegraphics[width=0.6\linewidth]{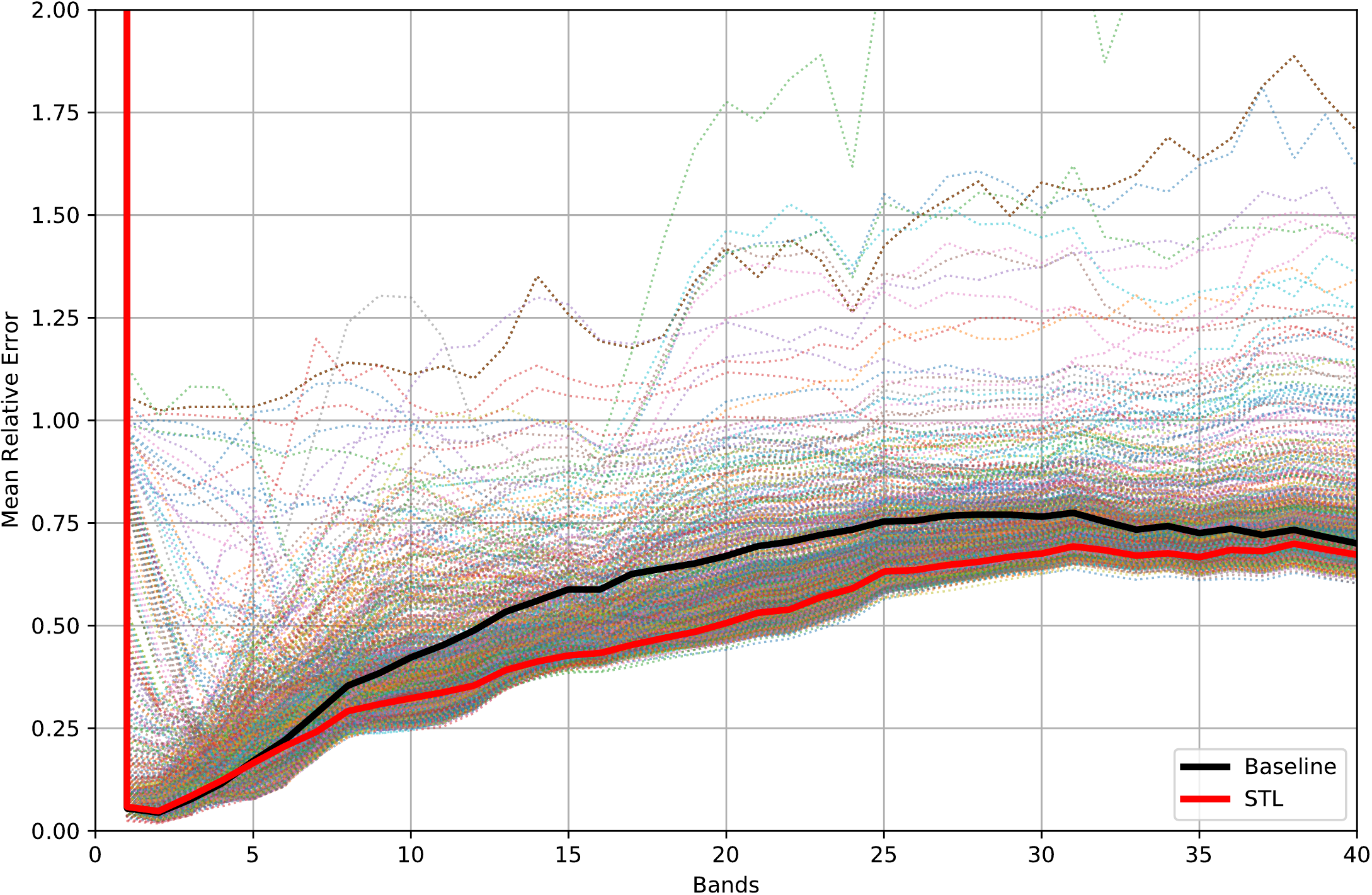}
	\caption{MRE of a total of 1482 runs with Fourier weights chosen by a grid and random search around the parameters of STL. \label{fig:search}}
	\vspace{-30pt}
\end{figure}

For a quantitative analysis we use the mean relative error (MRE) of the norm that is defined as the sum of differences in complex norm between the reconstructed and the ground truth samples, normalized by the sum of magnitudes of the ground truth.
Figure~\ref{fig:search} shows the MRE for different bands for the baseline model \cite{Kim2019} (black) and the frequency-aware STL (red). It can be seen that the largest improvement of up to 10\% is achieved for mid-frequency bands (15-25), while for low and high frequencies the resulting MRE is similar to the baseline. 

We performed an extensive grid and random search around the parameters of STL. In the grid search each parameter was set to 50\%, 100\%, 150\%, and 200\% of its STL value, except for the highest band which did not include 150\%. For the random search we sampled from the standard normal distribution for each parameter. If the resulting value was negative we divided the original STL value by the magnitude, otherwise we multiplied. For better comparison we normalized the final list of values for each run. 
Figure~\ref{fig:search} shows the resulting MRE plots for 1482 runs (768 grid search, 714 random search). Although we tested a wide range of different parameter combinations, none of them seem to significantly outperform STL over the whole range of frequencies. 

\subsection{Reconstruction Quality}

\begin{figure}[t!]
	\centering
	\newcommand*{\width}{0.25}
	\includegraphics[width=\width\linewidth, trim= 5 20 19 20, clip]{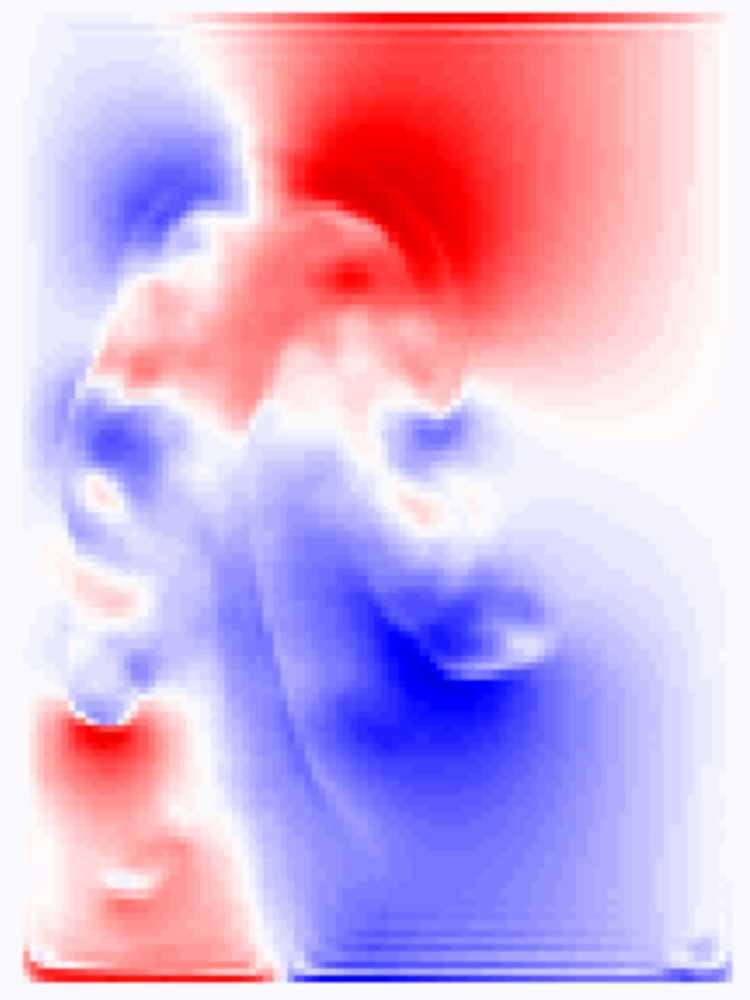}
	\includegraphics[width=\width\linewidth, trim= 5 20 19 20, clip]{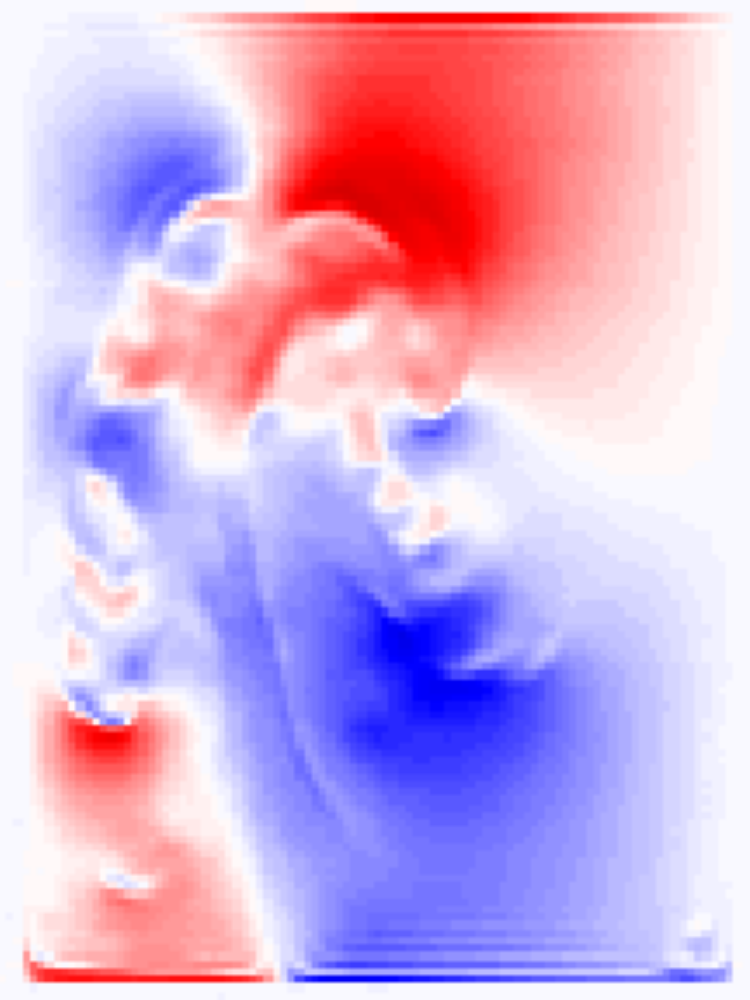}
	\includegraphics[width=\width\linewidth, trim= 5 20 19 20, clip]{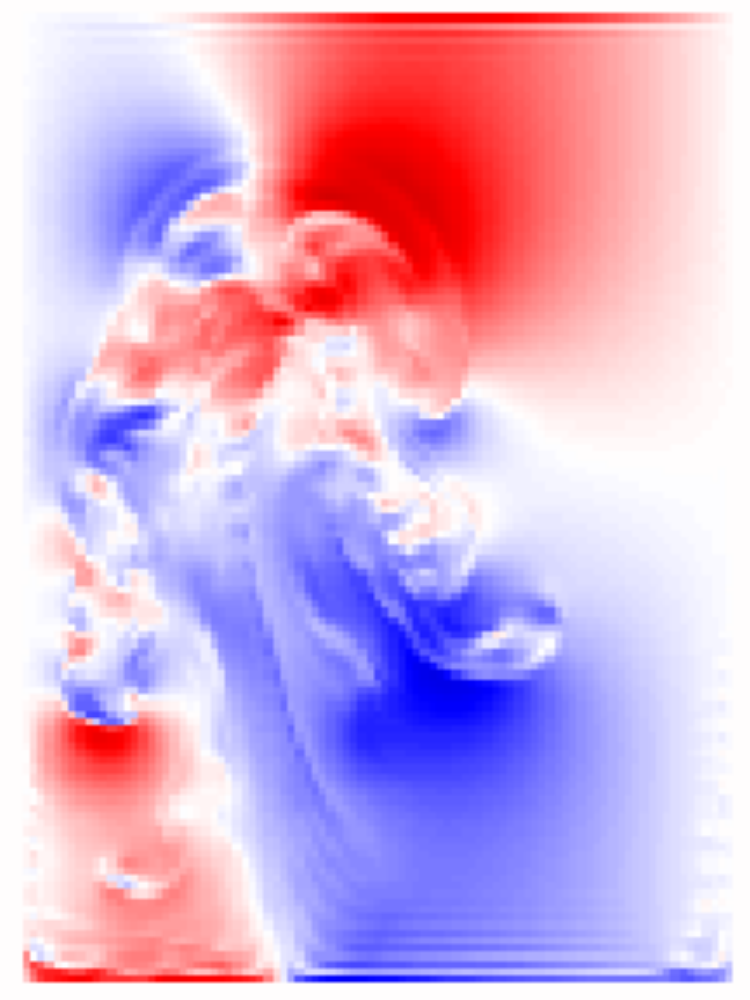}
	\includegraphics[width=\width\linewidth, trim= 12 20 12 20, clip]{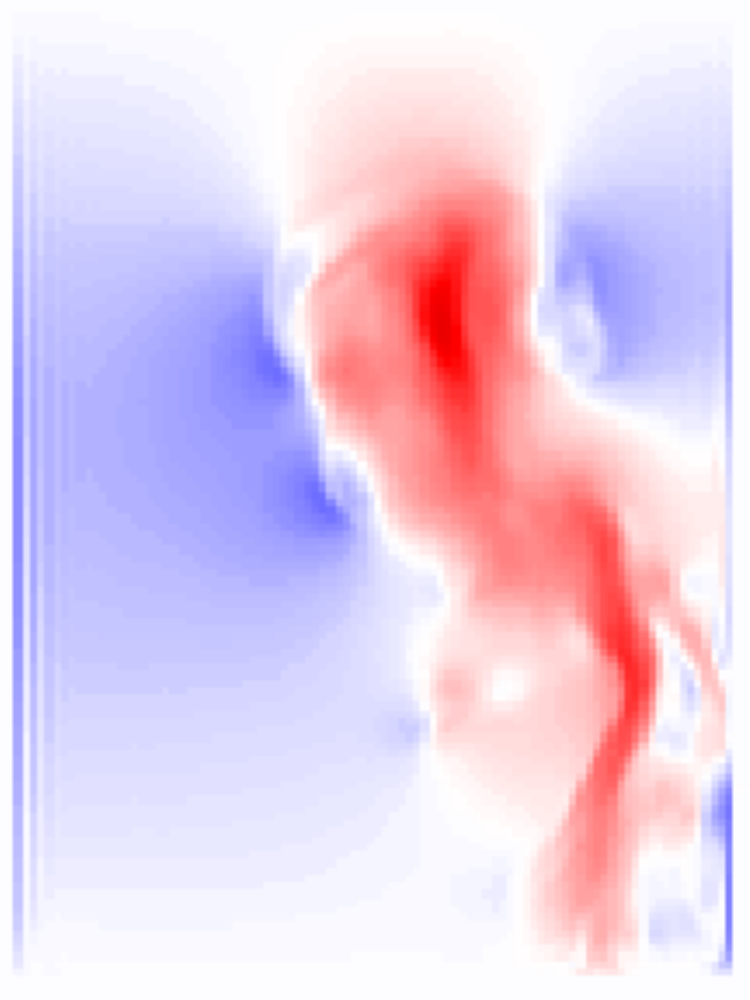}
	\includegraphics[width=\width\linewidth, trim= 12 20 12 20, clip]{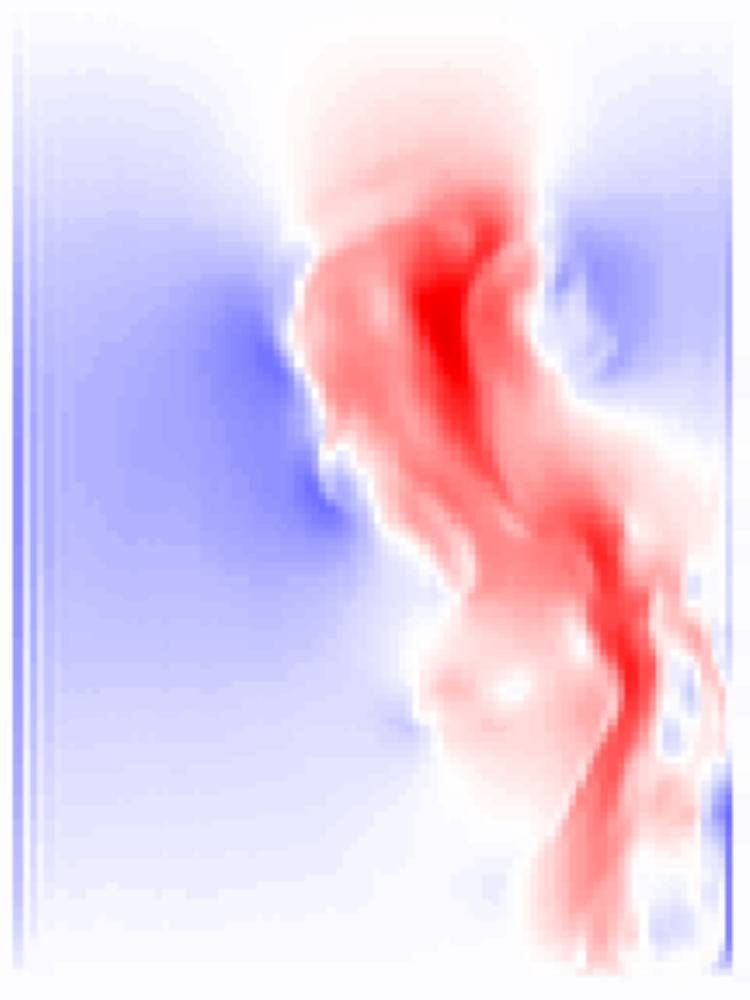}
	\includegraphics[width=\width\linewidth, trim= 12 20 12 20, clip]{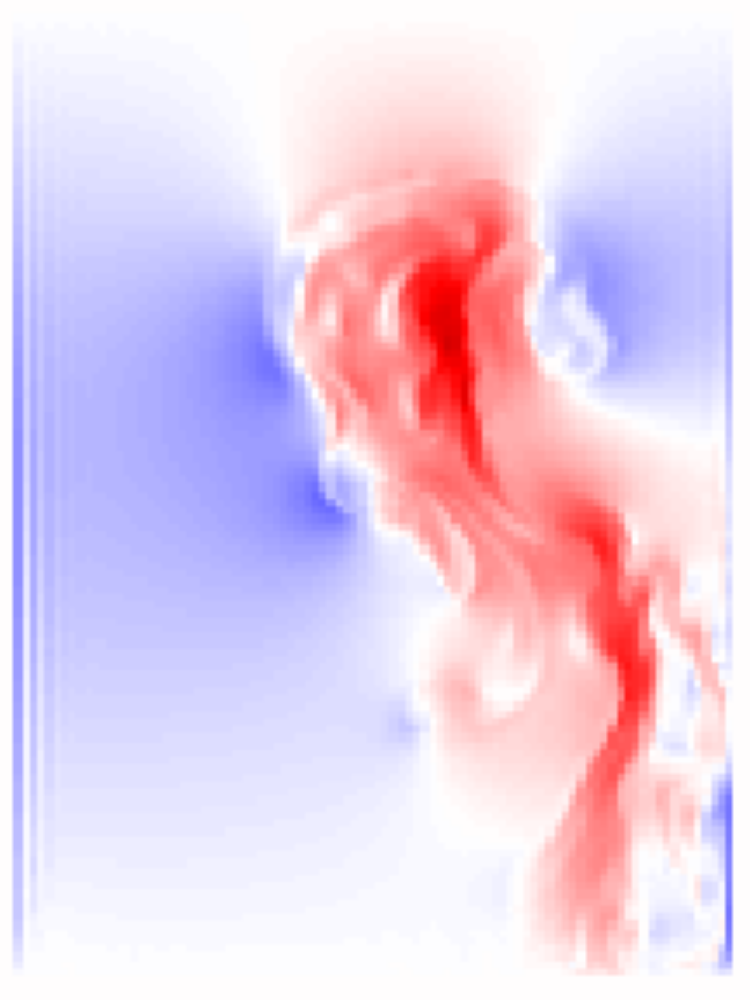}
	\includegraphics[width=\width\linewidth, trim= 12 20 12 20, clip]{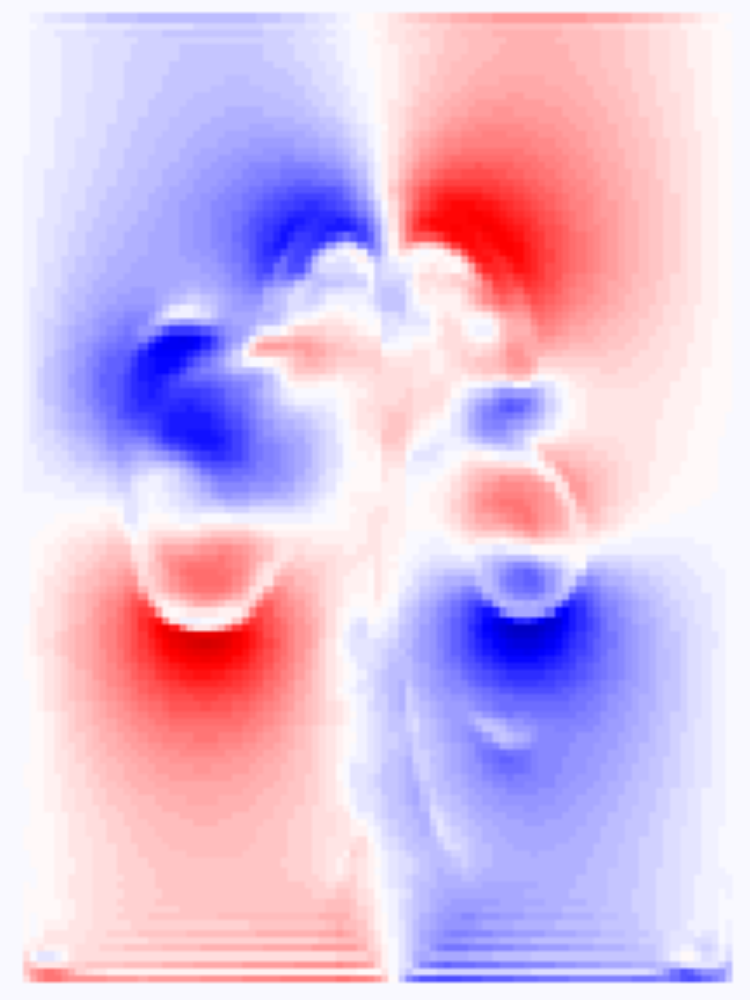}
	\includegraphics[width=\width\linewidth, trim= 12 20 12 20, clip]{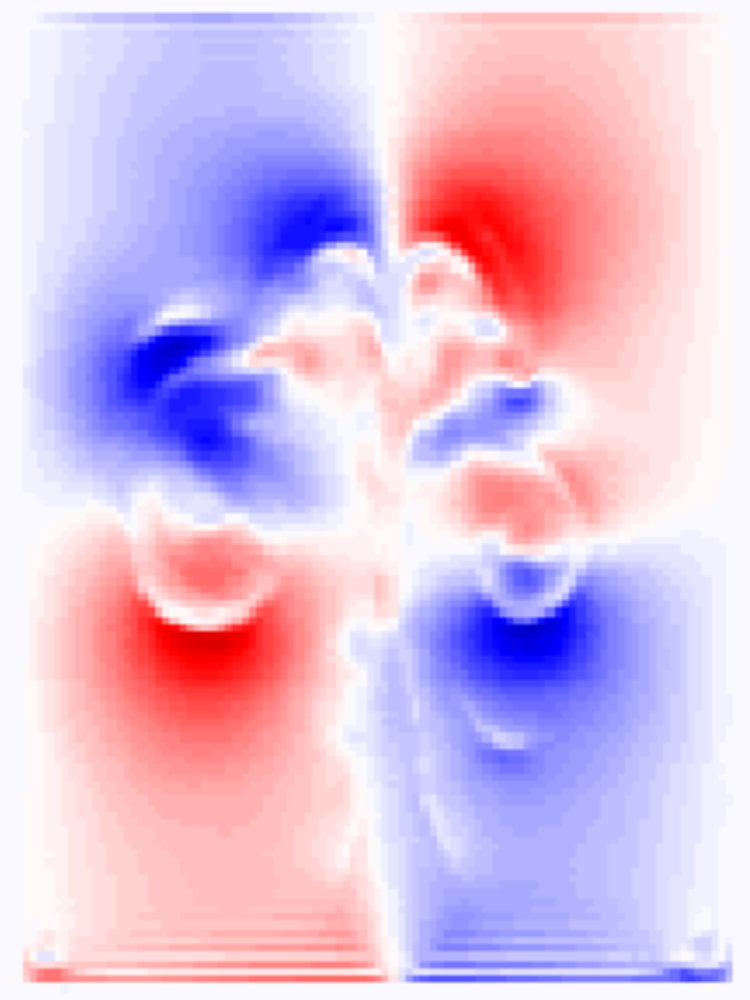}
	\includegraphics[width=\width\linewidth, trim= 12 20 12 20, clip]{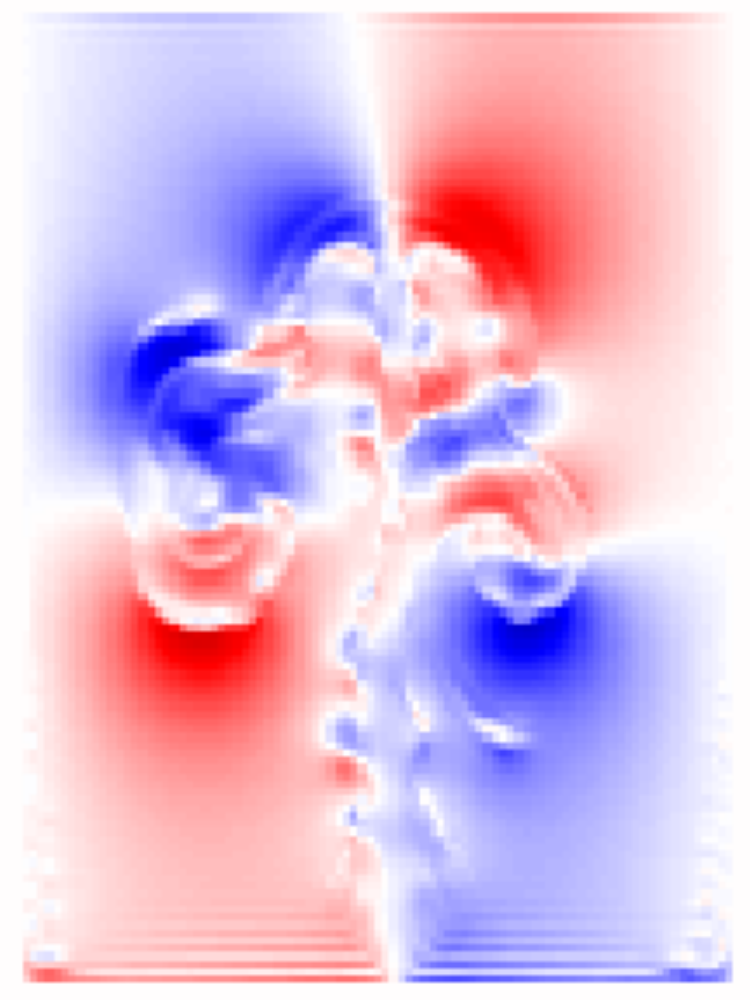}
	\caption{Velocity fields from baseline (left), STL (middle), and ground truth (right).}
\label{fig:detail}
\vspace{-12pt}
\end{figure}

\begin{figure} [t!]
	\centering
	\newcommand*{\width}{0.275}
	\includegraphics[width=\width\linewidth, trim= 20 0 100 140, clip]{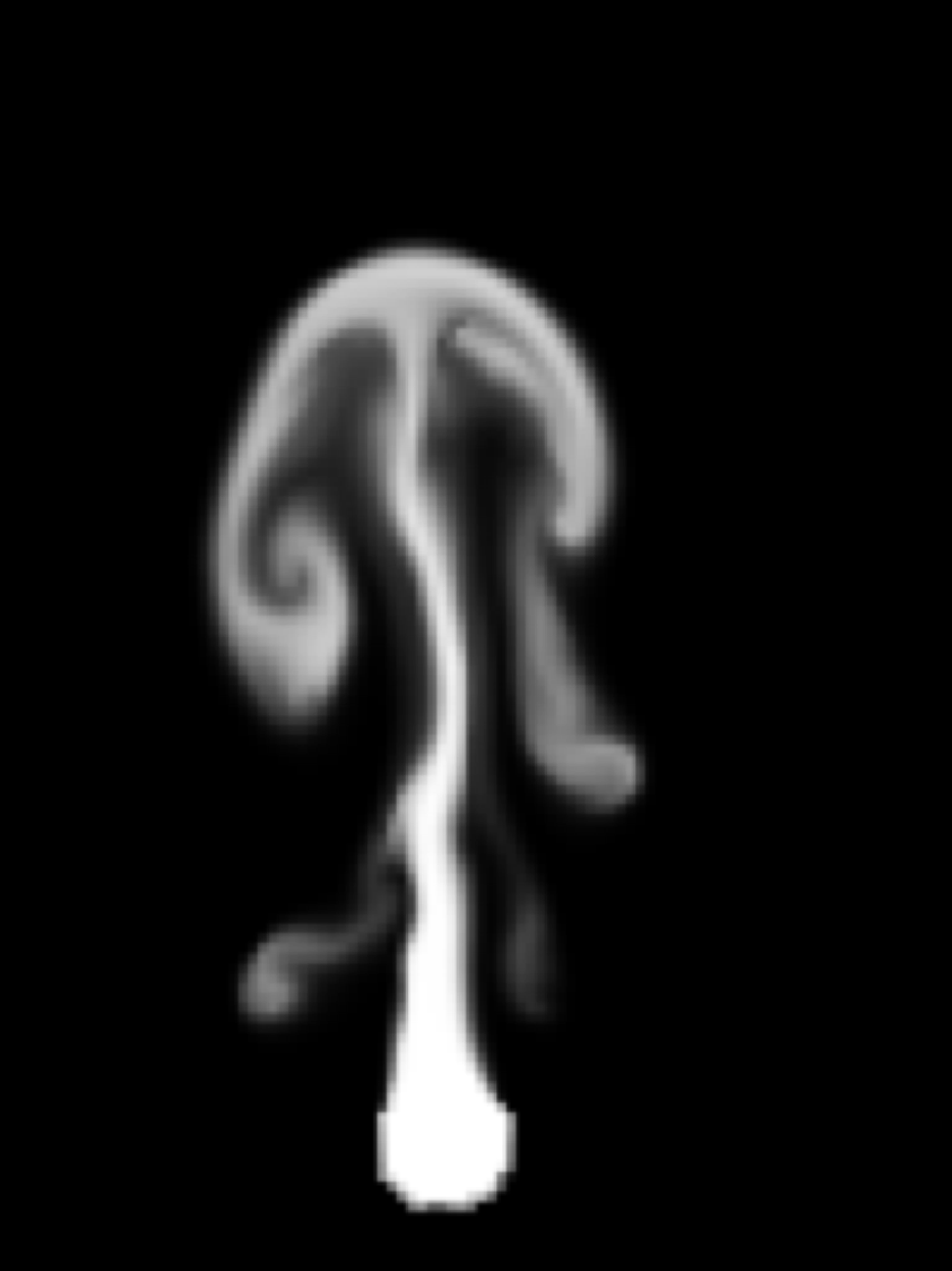}
	\includegraphics[width=\width\linewidth, trim= 20 0 100 140, clip]{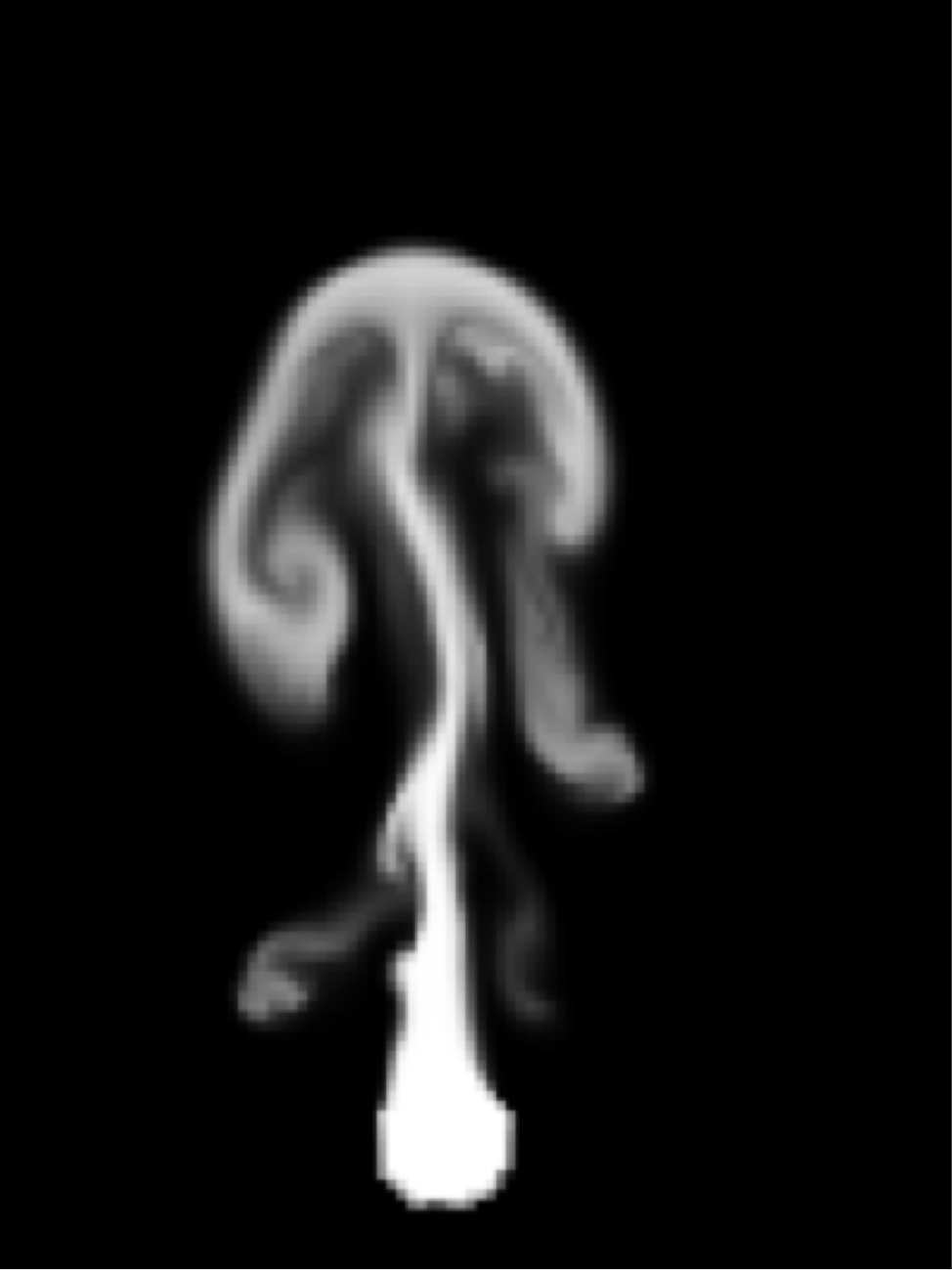}
	\includegraphics[width=\width\linewidth, trim= 20 0 100 140, clip]{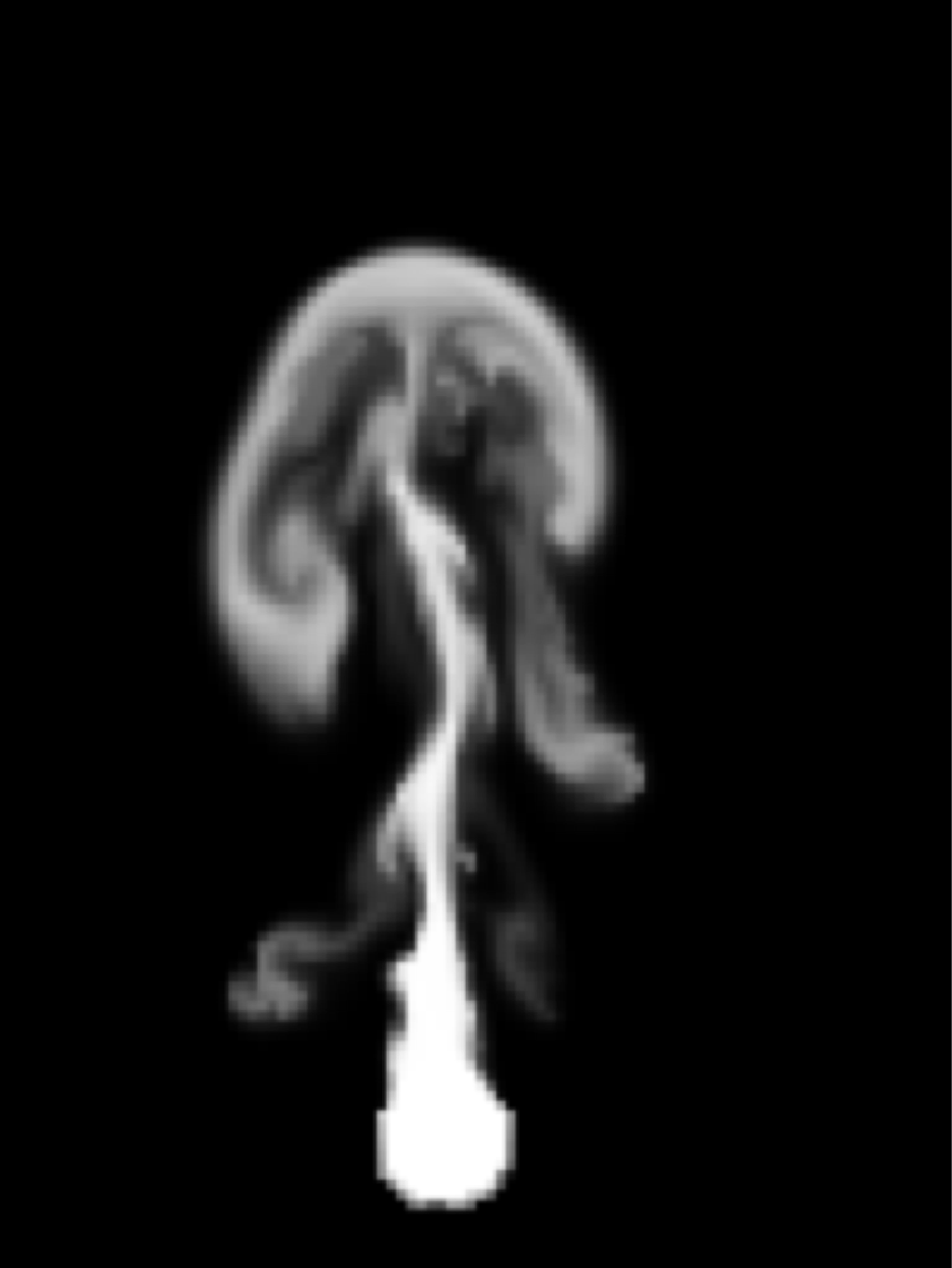}
	\caption{Advected densities with baseline (left), STL (middle), and ground truth (right).}
	\label{fig:simulation}
	\vspace{-12pt}
\end{figure}

The lower error in the mid-frequency bands improves the reconstructed velocity field as shown in Figure~\ref{fig:detail} (from left to right: baseline, STL and ground truth) and accordingly the resulting density field that is advected with the reconstructed velocity in Figure~\ref{fig:simulation}. 
Although some fine structures are better captured with STL than with the baseline, there is still a discrepancy between the reconstructions and the ground truth. 
%
%
%

\subsection{Comparison with GAN}
\begin{figure}[h]
	\centering
	\newcommand*{\width}{0.26}
	\includegraphics[width=\width\linewidth, trim= 50 70 10 80, clip]{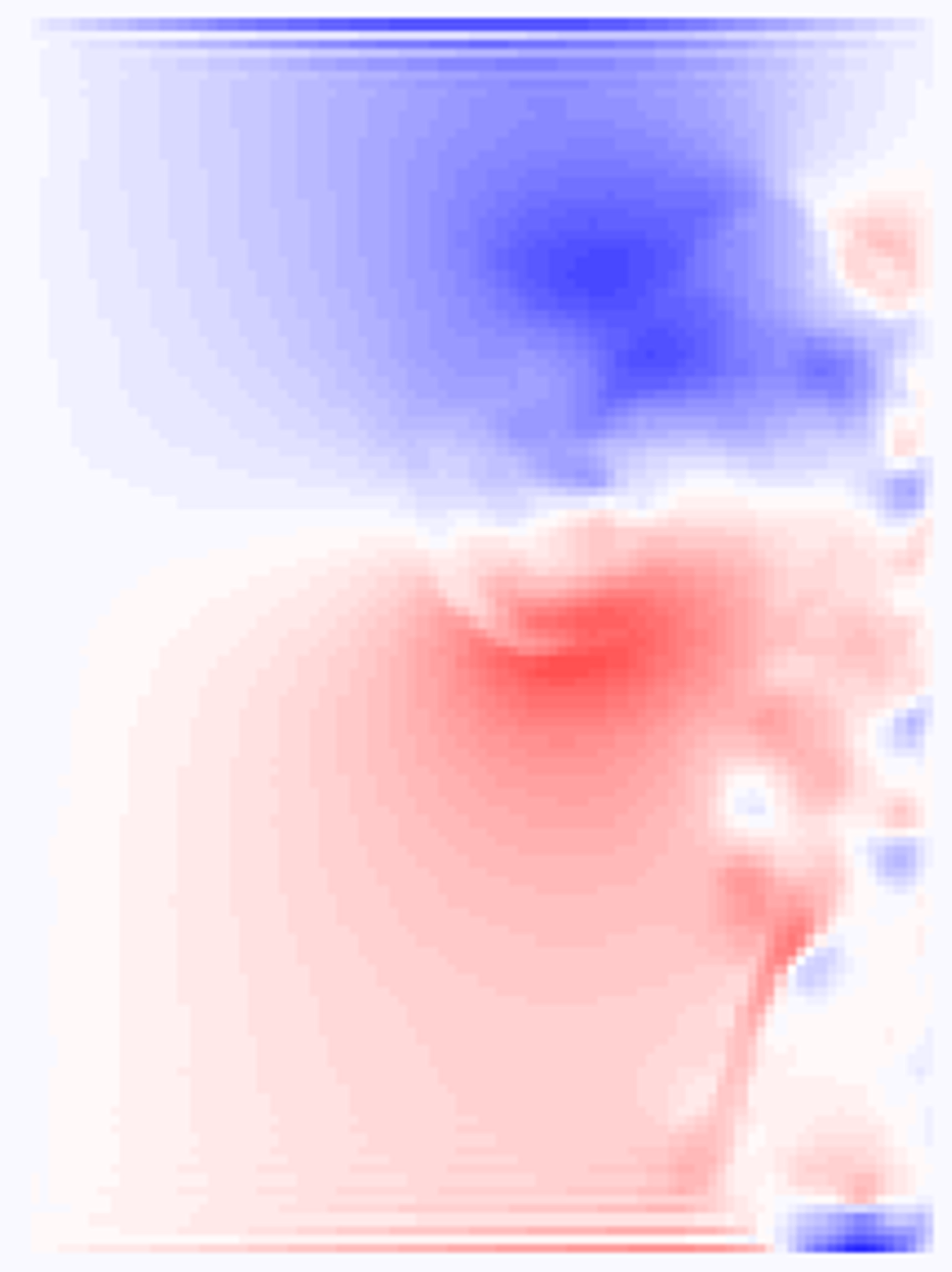}
	\includegraphics[width=\width\linewidth, trim= 50 70 10 80, clip]{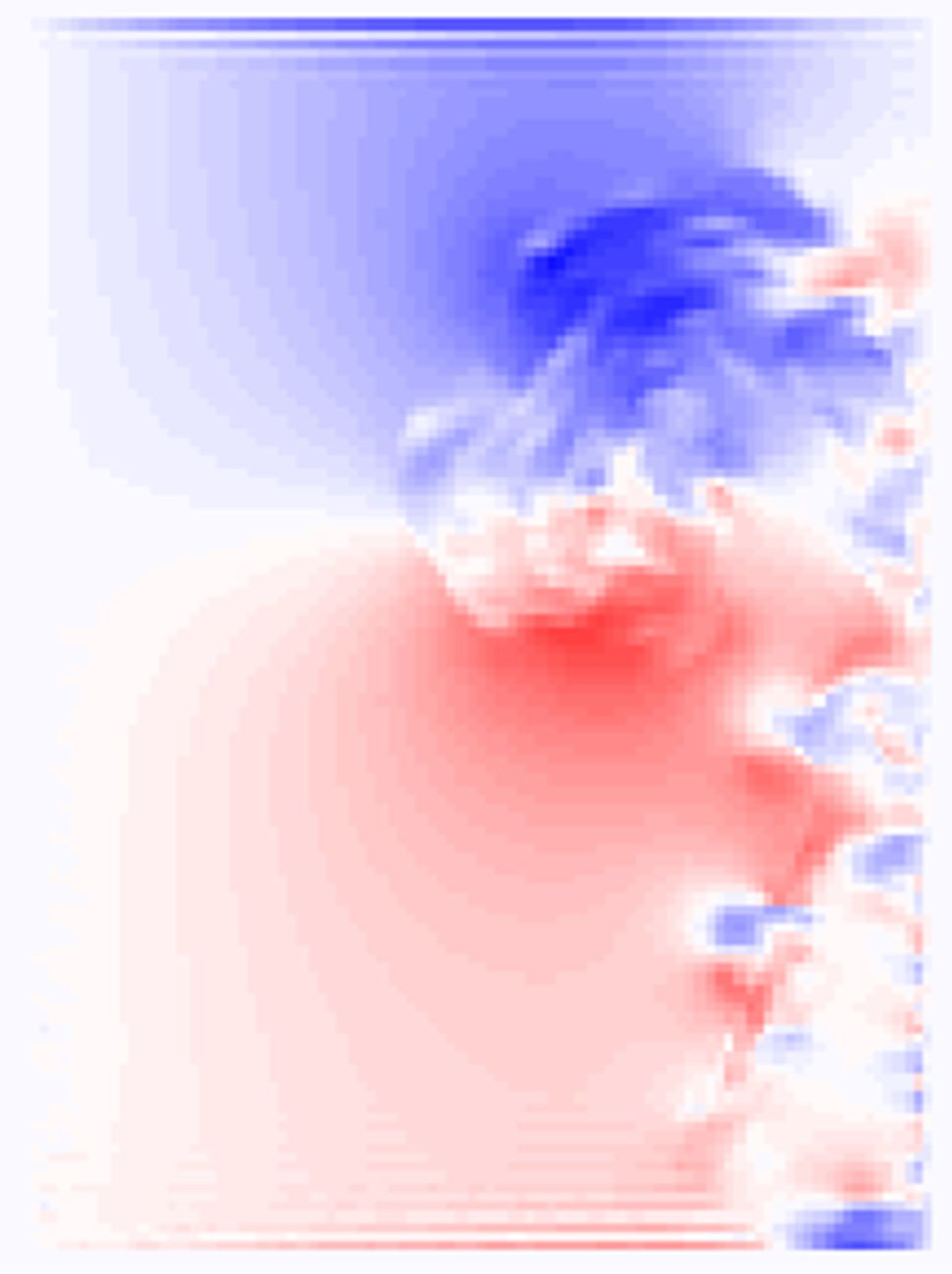}
	\includegraphics[width=\width\linewidth, trim= 50 70 10 80, clip]{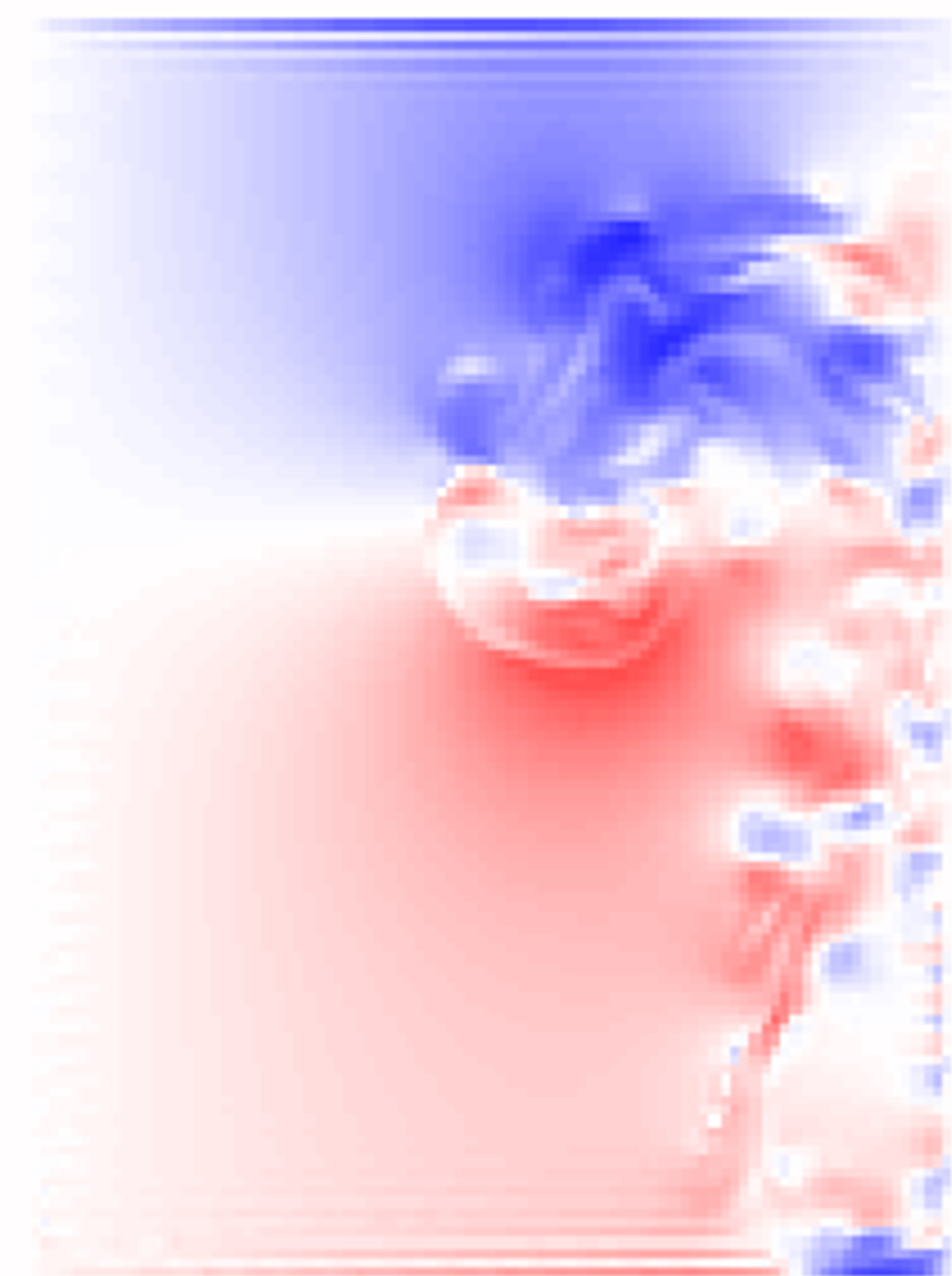}
	\caption{Visual comparison of the baseline (left), GAN (middle) and ground truth (right). \label{fig:gan_results}}
	\vspace{-15pt}
\end{figure}

\begin{figure}[h]
\centering
	\includegraphics[width=0.47\linewidth, trim= 0 0 0 0, clip]{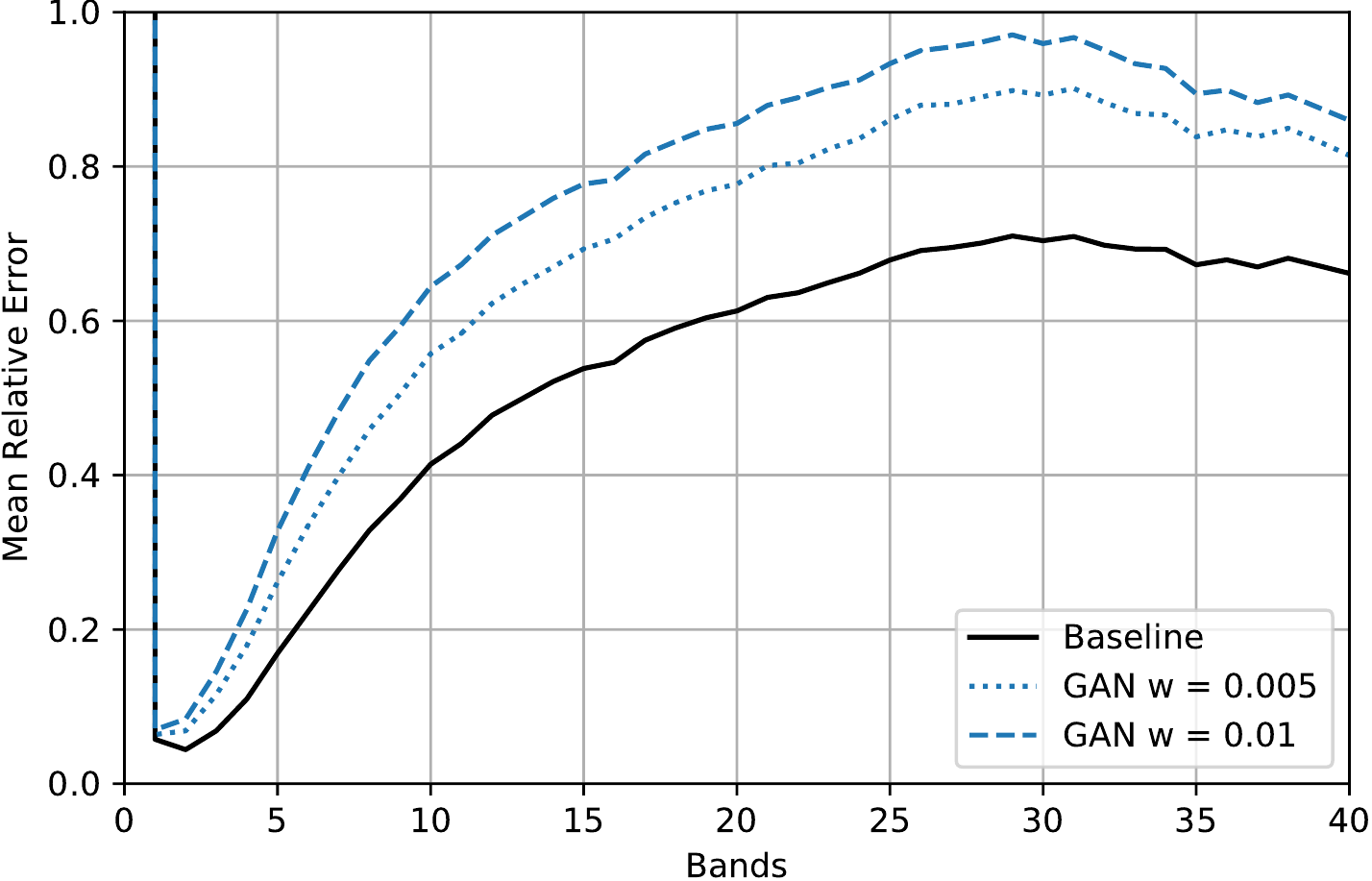}
	\includegraphics[width=0.49\linewidth, trim= 0 0 0 10, clip]{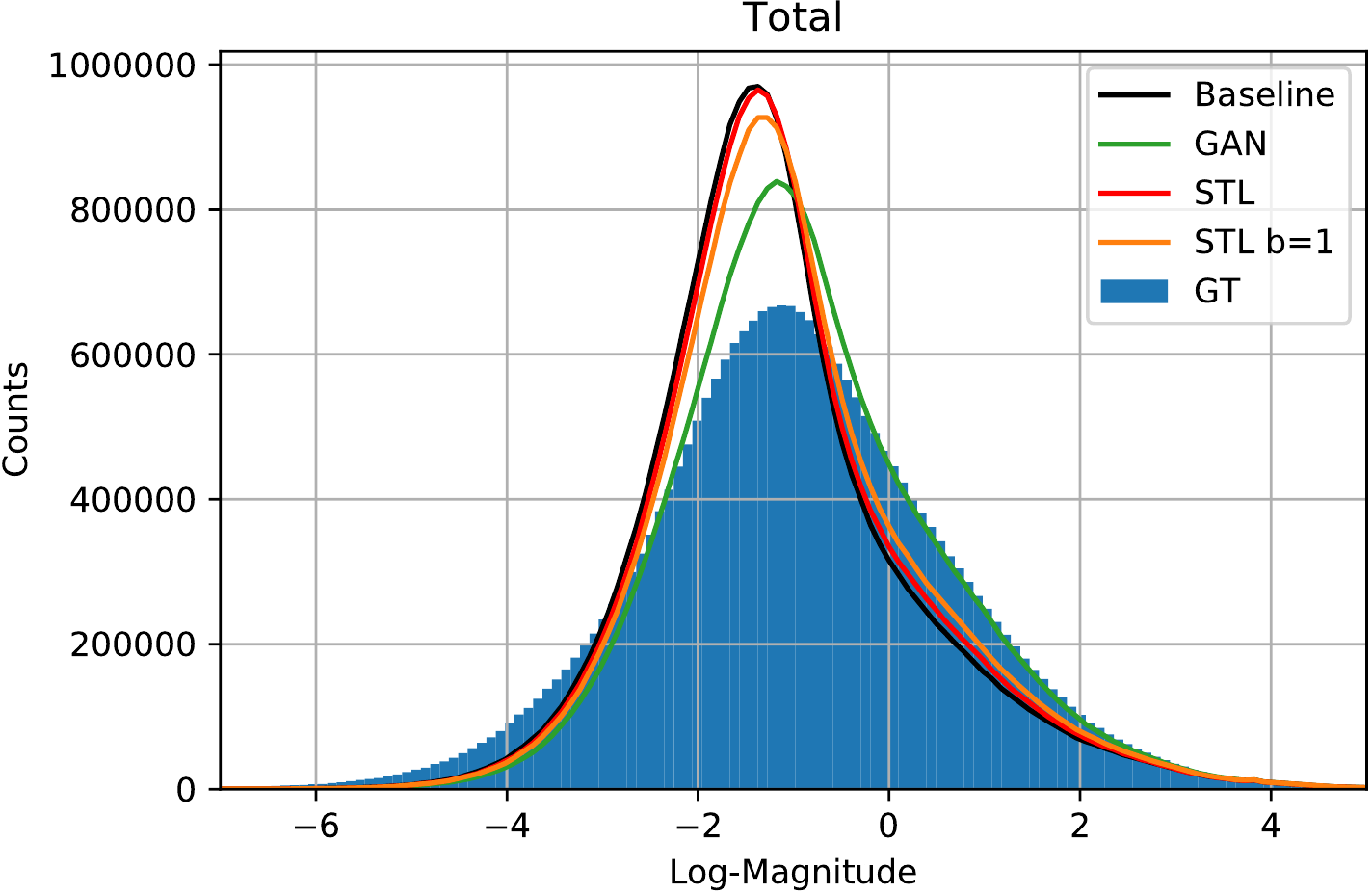}
	\caption{Left: MRE of the baseline (black) and GAN (blue). Right: Histogram of the log-magnitudes for different training methods and the ground truth distribution.\label{fig:gan_mre}}
	\vspace{-16pt}
\end{figure}

We compared our approach with a generative adversarial network (GAN), as GANs are known to perform well for reconstructing details in images. We used PatchGAN~\cite{Isola2016} in our implementation. Figure~\ref{fig:gan_results} shows that the GAN can generate impressively detailed flow structures that resemble the ground truth data. However, these generated structures only mimic flow details and in fact are less physical than the corresponding result with the baseline approach. This is also reflected by the higher MRE of GANs compared to the baseline, which is shown in Figure~\ref{fig:gan_mre} (left).
Moreover, we observed that GANs are more sensitive to parameters and have an increased training time compared to our STL approach, which is comparable to the baseline performance.
%
%

\subsection{Histogram Analysis}
When investigating the histogram of the log-magnitudes of the Fourier transform in Figure~\ref{fig:gan_mre} (right), one can notice the ability of the GAN training to better match the underlying distribution since the discriminator learns to distinguish features that are not present on the generated data. 
In contrast, STL moves the curve only slighly towards the ground truth distribution. Due to this observation, we also evaluated a histogram loss~\cite{Risser2017}, but we found large discrepancies between the resulting velocity field and ground truth, even if histograms better match. We presume that matching a global histogram of the magnitudes does not capture the characteristics of flow data well, and also that the interplay between magnitude and phase might need to be considered in the optimization.
%
%

\section{Conclusion}
\label{sec:Conclusions}

Our evaluation has shown that the baseline method does not minimize the error efficiently for higher frequency bands of the input data, and that a loss function is needed that can better discriminate between low and high frequencies of the input. Results indicate that the inclusion of spectral approaches is a promising direction to improve the reconstruction quality in mid-frequency bands.
However, reconstructions still deviate from the ground truth.
More research is needed to evaluate the shape and number of bands used in the Fourier loss to develop a definite conclusion on the strengths and weaknesses of the proposed loss. Future work could also include the evaluation of perceptual losses and wavelet approaches. An interesting finding is also that generative adversarial networks can better approximate the ground truth histogram distribution of the Fourier transform's log-magnitudes and impressively improve the perceived quality. 
However, the reconstructed flow is non-physical, leading to higher MREs in all bands compared to the baseline and preventing applications related to flow data compression.

\section{Acknowledgments}
\label{sec:Acknowledgments}
This work was supported by the Swiss National Science Foundation (Grant No. 200021\_168997).

\bibliographystyle{eg-alpha-doi}  
\bibliography{frequency}        


\end{document}